\documentclass[10pt,twocolumn,letterpaper]{article}
\usepackage{cvpr}
\usepackage{pifont}
\usepackage{amsmath}
\usepackage{float}
\usepackage{mathtools}

\definecolor{cvprblue}{rgb}{0.21,0.49,0.74}
\usepackage[pagebackref,breaklinks,colorlinks,allcolors=cvprblue,hypertexnames=false]{hyperref}

\title{Layered Motion Fusion: Lifting Motion Segmentation to 3D in Egocentric Videos}

\author{
Vadim Tschernezki$^{1,2}$ ~ ~ Diane Larlus$^2$ ~ ~ Iro Laina$^1$ ~ ~ Andrea Vedaldi$^1$\\
\centering
\begin{minipage}{.4\textwidth}
\centering
\small{~\\ $^1$Visual Geometry Group\\University of Oxford\\}
{\tt\small \{vadim,iro,vedaldi\}@robots.ox.ac.uk}
\end{minipage}
\begin{minipage}{.4\textwidth}
\centering
\small{~\\~\\ $^2$NAVER LABS Europe\\}
{\tt\small diane.larlus@naverlabs.com}
\end{minipage}
}

\begin{document}
\newcommand{\cmark}{\ding{51}}%
\newcommand{\xmark}{\ding{55}}%
\maketitle
\begin{abstract}
Computer vision is largely based on 2D techniques, with 3D vision still relegated to a relatively narrow subset of applications. However, by building on recent advances in 3D models such as neural radiance fields, some authors have shown that 3D techniques can at last improve outputs extracted from independent 2D views, by fusing them into 3D and denoising them. This is particularly helpful in egocentric videos, where the camera motion is significant, but only under the assumption that the scene itself is static. In fact, as shown in the recent analysis conducted by EPIC Fields, 3D techniques are ineffective when it comes to studying dynamic phenomena, and, in particular, when segmenting moving objects. In this paper, we look into this issue in more detail. First, we propose to improve dynamic segmentation in 3D by fusing motion segmentation predictions from a 2D-based model into layered radiance fields (Layered Motion Fusion).  However, the high complexity of long, dynamic videos makes it challenging to capture the underlying geometric structure, and, as a result, hinders the fusion of motion cues into the (incomplete) scene geometry. We address this issue through test-time refinement, which helps the model to focus on specific frames, thereby reducing the data complexity. This results in a synergy between motion fusion and the refinement, and in turn leads to segmentation predictions of the 3D model that surpass the 2D baseline by a large margin. This demonstrates that 3D techniques can enhance 2D analysis even for dynamic phenomena in a challenging and realistic setting.
\end{abstract}

\section{Introduction}%
\label{sec:intro}

Forty years ago, pioneers like Marr~\cite{marr82vision:} argued that 3D representation should be a foundation of computer vision.
However, the field has since evolved %
differently, reducing image and video understanding to 2D pattern recognition, first with the introduction of visual representations like bags of visual words~\cite{lowe99object,csurka04visual,sivic06video,perronnin06fisher} and then learned ones~\cite{krizhevsky12imagenet,simonyan15very,he15deep,girshick15fast,vaswani17attention,dosovitskiy21an-image}.

Recently, several authors have shown that 3D representations can, if not replace, at least enhance 2D techniques.
An approach is to \emph{fuse} 2D information extracted from individual views of a scene into a coherent 3D reconstruction.
Examples include methods like Semantic NeRF~\cite{zhi21in-place}, 
N3F~\cite{tschernezki22neural}, 
DFF \cite{kobayashi2022distilledfeaturefields}, LERF~\cite{lerf2023}, GARField~\cite{kim24garfield:}.
These techniques take various types of 2D outputs (\eg, semantic or instance segmentation, image or text features) and project them into a coherent 3D reconstruction while also removing noise and compressing information.

While these methods show that 3D \emph{can} improve 2D processing, so far this has been demonstrated only 
in restricted settings;
most of these methods require multi-view images of a \emph{static} scene to work. 
However, in real applications, the scene itself is often dynamic, and, arguably, the dynamic part is often the most interesting one.
Thus, 3D techniques need to be able to handle dynamic content in order to be useful in practice.
As 3D methods mature, however, we can expect them to become useful in these more challenging scenarios as well, particularly when the scene content is dynamic \emph{and} the camera moves significantly.
A great example of such a scenario is egocentric videos, in which a camera is attached to a person as they navigate and interact with their environment, resulting in camera movement that mirrors their actions within the scene.

A few authors have explored this challenging scenario in the broader context of monocular video understanding~\cite{tschernezki21neuraldiff, liang2023semantic, wu2022d}.
Out of those, NeuralDiff~\cite{tschernezki21neuraldiff} focuses on monocular egocentric videos.
It uses an architecture based on neural radiance fields (NeRF)~\cite{mildenhall20nerf:} to decompose these videos into three layers:
a static background %
and two foreground layers, respectively modeling the semi-static part %
(objects that are currently stationary, but move at some point in the video) and the dynamic part (objects that presently move) of the scene. %
More recently, EPIC Fields~\cite{tschernezki23epic} conducted a systematic study of 3D techniques for long egocentric videos, introducing a large annotated dataset and performing comprehensive empirical evaluations of existing techniques for novel view synthesis and scene decomposition into static, semi-static and dynamic objects. One of the main findings of that study is that \textit{the performance of 3D methods lags behind that of the 2D baseline in terms of segmenting dynamic components}.

Inspired by recent developments in the distillation of features \cite{kobayashi2022distilledfeaturefields, tschernezki22neural,lerf2023} and the fusion of labels \cite{zhi21in-place,siddiqui2023panoptic,kundu2022panoptic} into 3D representations, we aim to improve the 3D segmentation by fusing the motion of a 2D-based method into a 3D representation. This leads us to our key question: \textit{can a 3D representation improve the performance of a 2D neural network in understanding dynamic phenomena?}
To answer it, we consider a strong unsupervised 2D-based motion segmentation method, Motion Grouping (MG)~\cite{yang2021self}, and suggest to fuse its predictions into our \emph{dynamic} 3D model. We observe that while such models capture only parts of the dynamic objects (making them incomplete), they are precise and therefore suitable for being fused into a 3D representation as they are similar to ``sparse'' (incomplete, but precise) labels as in Semantic-NeRF~\cite{zhi21in-place}.
To do so, we develop a new \emph{motion fusion} technique based on a layered representation of dynamic NeRFs. We first show that the fusion of the motion segmentation predictions into the dynamic layer already results in significant improvements of the segmentation capability. Additionally, we find that we can further improve the 3D model by regularizing the semi-static layer with the \textit{same} segmentation predictions from the 2D-based model. We achieve this by penalizing the semi-static layer from predicting anything that the motion segmentation model thinks is dynamic. %
These constraints work in synergy to enhance the overall performance of the model.
Because we fuse motion into both the semi-static and dynamic layers, we refer to this method as Layered Motion Fusion (LMF).

While the fusion of motion improves segmentation significantly, we find that this process is still limited by the high data complexity of egocentric videos. Specifically, we observe that 3D-based models can only fuse motion segmentation predictions into the geometry as they have learned it from the scene. If the scene's geometry is too complex\,---\,such as in long, complex egocentric videos\,---\,the model fails to capture it and, in turn, cannot fuse motion into the (missing) geometry.
To address this issue, we suggest considering a setting where test-time adaptation~\cite{liang2023comprehensive} and test-time refinement~\cite{izquierdo2023sfm} can be applied.
We show that fine-tuning the model to the subset of frames we wish to analyze enhances the model's ability to capture the scene's geometry, thereby allowing it to fuse motion more accurately.

In summary, our contributions are as follows:
(1) We propose a new motion fusion technique for layered radiance fields that boosts the segmentation of dynamic objects in egocentric videos by a large margin. Additionally, we observe improvements in the segmentation of semi-static objects, highlighting the benefits of a layered fusion. To our knowledge, this represents the first attempt at fusing motion segmentation into dynamic radiance fields.
(2) We propose test time refinement to further boost segmentation performance by focusing the optimization on selected frames to reduce scene complexity.
(3) We solve the issue observed in \cite{tschernezki23epic} of the inferior results of 3D models compared to 2D ones for unsupervised dynamic object segmentation, and show that our proposed method outperforms all results reported in~\cite{tschernezki23epic}. %
This underscores the potential of 3D computer vision techniques to \emph{enhance} the performance of 2D video understanding methods, even in challenging, highly dynamic scenarios. We hope that this finding will encourage others %
to explore 3D vision for understanding this data, where we believe it has considerable potential.

\section{Related work}
\label{sec:rw}

\paragraph{NeRF and dynamics.} Neural Radiance Fields (NeRFs) were introduced in \cite{mildenhall20nerf:} as a way to synthesize novel views in 3D scenes. 
Initially restricted to static scenes, several methods have extended NeRF to dynamic scenes.
There are two main approaches.
The first one adds time as an additional dimension to radiance fields~\cite{martinbrualla2020nerfw,tschernezki21neuraldiff,xian21space-time,gao21dynamic,wang22fourier,li22neural,kplanes_2023,Cao2023HEXPLANE}. The other one 
explicitly learns a 3D flow and reduces the reconstruction to a canonical (static) one~\cite{pumarola2021d,yoon20novel,park2021nerfies,wang21neural,tretschk21non-rigid,li21neural,du21neural,yuan2021star,song22nerfplayer:,guo22neural,fang22fast,li2023dynibar,liu2022devrf}. As noted in \cite{tschernezki23epic}, dynamic neural rendering has been mostly applied to synthetic or simple environments (small camera baseline and sequence lengths up to 60 seconds). 
To encourage the use of dynamic neural rendering in more complex and realistic environments, the EPIC Fields benchmark \cite{tschernezki23epic} was proposed. It consists of long and complex egocentric video sequences and is associated with difficult scene understanding tasks. This benchmark highlights the unresolved issues of recent NeRF-like methods in rendering dynamic parts of scenes in long videos. Our work explicitly addresses these challenges.

\paragraph{NeRF and semantics.} 
Other research has focused on the potential of neural rendering to enhance the semantic understanding of the 3D model of a scene.
For example, Semantic NeRF~\cite{zhi21in-place,vora21nesf:} fuses semantic labels with static scenes. Others
explore panoptic segmentation~\cite{kundu2022panoptic,fu22panoptic,siddiqui2023panoptic,wang2022dm,bhalgat23contrastive},
which extends semantic segmentation with the ability to differentiate between instances of the same class. Besides integrating labels into a 3D scene representation, a related line of work \cite{tschernezki22neural,kobayashi2022distilledfeaturefields,liang2023semantic,lerf2023} has proposed to enhance NeRFs with a separate prediction head that captures semantic \textit{features} from pre-trained ViTs~\cite{caron21emerging,li2022languagedriven,touvron21training} or vision-language features from CLIP~\cite{radford21learning}. This merges the open-world knowledge from 2D models into 3D scene representations and extends them to applications such as the retrieval or editing of objects inside of scenes~\cite{tschernezki22neural,kobayashi2022distilledfeaturefields}. The authors of \cite{yang2021self,zhang2023nerflets} even capture the semantics of a scene through the decomposition of objects of this scene with individual radiance fields. %
Other related methods \cite{fan2023nerfsos, xie2021fig, yu2022unsupervised, tschernezki21neuraldiff,sharma2023neural, ost2021neural, mirzaei2022laterf,wu2022d} 
rely on neural rendering techniques to distinguish between objects and backgrounds 
with no or minimal supervisory signals. We specifically focus on the segmentation of dynamic objects, which has only been explored by few works~\cite{tschernezki21neuraldiff, liang2023semantic, wu2022d}.

\paragraph{Distillation of 2D models into 3D.}
While most work related to `NeRF and semantics' directly use labels, another option is to integrate semantic knowledge through the distillation of models designed to receive 2D images as input into 3D representations. This idea has been explored already before NeRF and is known as multi-view semantic fusion methods~\cite{hermans2014dense, mccormac2017semanticfusion, sunderhauf2017meaningful, ma17multi-view, mascaro21diffuser:,vineet2015incremental}. Similar to approaches such as
Semantic NeRF~\cite{zhi21in-place} and Panoptic Lifting~\cite{siddiqui2023panoptic}, these methods combine multiple, potentially noisy or partial, 2D semantic observations
and re-render them to obtain clean and multi-view consistent labels. Other research incrementally builds semantic maps using SLAM~\cite{kundu2014joint,tateno2017cnn,narita2019panopticfusion}.
The advent of NeRF has increased the application of distillation for 3D representations as shown in DFF~\cite{kobayashi2022distilledfeaturefields},
N3F~\cite{tschernezki22neural}, and LERF~\cite{lerf2023}. These methods apply 3D fusion directly to supervised and unsupervised dense features in order to transfer semantics into the 3D space. This benefits applications such as scene editing, object retrieval and zero-shot 3D segmentation. More recent approaches such as \cite{zuo2024fmgs,bhalgat2024n2f2,qin2024langsplat,zhou2024feature} extend these ideas to the more efficient Gaussian Splatting~\cite{kerbl3Dgaussians} rendering technique that significantly speeds up training and rendering,
which in turn makes the fusion of labels/features more easily applicable. Similarly, we distill 2D knowledge from a model, but with the difference that our 2D model is specialized in motion segmentation. 

\paragraph{Motion and object segmentation.} As shown in \cite{tschernezki21neuraldiff}, objects can be segmented in dynamic videos without supervision by combining motion cues with a NeRF-like architecture. 
This can also be done using standard 2D approaches such as background subtraction \cite{bouwmans2014traditional} or motion segmentation \cite{08motionseg, 21motionseg, OB14b}. 
The latter typically requires optical flow, which is subject to ambiguities~\cite{cvpr20motionseg} and only reasons locally. 
Such approaches are also prone to errors in the presence of occlusions or if dynamic objects temporarily remain static \cite{08motionseg, 21motionseg, cvpr21motionseg}. The mechanism behind motion segmentation has been extended to the segmentation of specific objects~\cite{bideau2016s, papazoglou2013fast, tokmakov2017learning, jain2017fusionseg, xie2019object}. In~\cite{bideau2016s} a probabilistic model acts upon optical flow to segment moving objects from the background. In~\cite{sundaram2010dense, brox2010object} pixel trajectories and spectral clustering are combined to produce motion segments, while \cite{matzen14scene} reconstructs urban scenes and discovers their dynamic elements such as billboards or street art by clustering 3D points in space and time. More recent work~\cite{wang2019zero, xie2019object,choudhury+karazija22gwm,lao2024divided,Lian_2023_CVPR,xie2022segmenting} such as Motion Grouping (MG)~\cite{yang2021self} combines classical motion segmentation with deep learning. Other examples include \cite{Lian_2023_CVPR}, which learns an image segmenter in the loop of approximating optical flow with constant segment flow and then refines it for more coherent appearance and statistical figure-ground relevance, and \cite{xie2022segmenting}, which segments moving objects via an object-centric layered representation.
Our method can be combined with these techniques
and refine them with dynamic neural rendering, resulting in cleaner and more accurate motion segmentation masks. 

\paragraph{Test-time refinement.} 
The complexity of egocentric videos can be approached by enabling the model to focus on specific parts of the 
video at test time.
Concretely, we are interested in improving the segmentation of \textit{specific frames} that the model receives as input. 
This is closely related to test-time adaptation~\cite{liang2023comprehensive}, which aims at adjusting a pre-trained model as test data becomes available. Predictions are made using the adjusted model.
Similarly, we can adapt 3D  
models to user selected frames and 
use their corresponding predicted motion segmentation as pseudo-labels. 
In the context of 3D vision, this paradigm is more specifically known as test-time refinement \cite{izquierdo2023sfm}. For example, SfM-TTR~\cite{izquierdo2023sfm} boosts the performance of single-view depth networks at test time using SfM multi-view cues, while \cite{chen2019self} learns depth, optical flow, camera pose and intrinsic parameters on a test sample in an online refinement setting. 

\section{Method}%
\label{sec:method}

\begin{figure*}[t]
\centering
\includegraphics[width=1.0\textwidth]{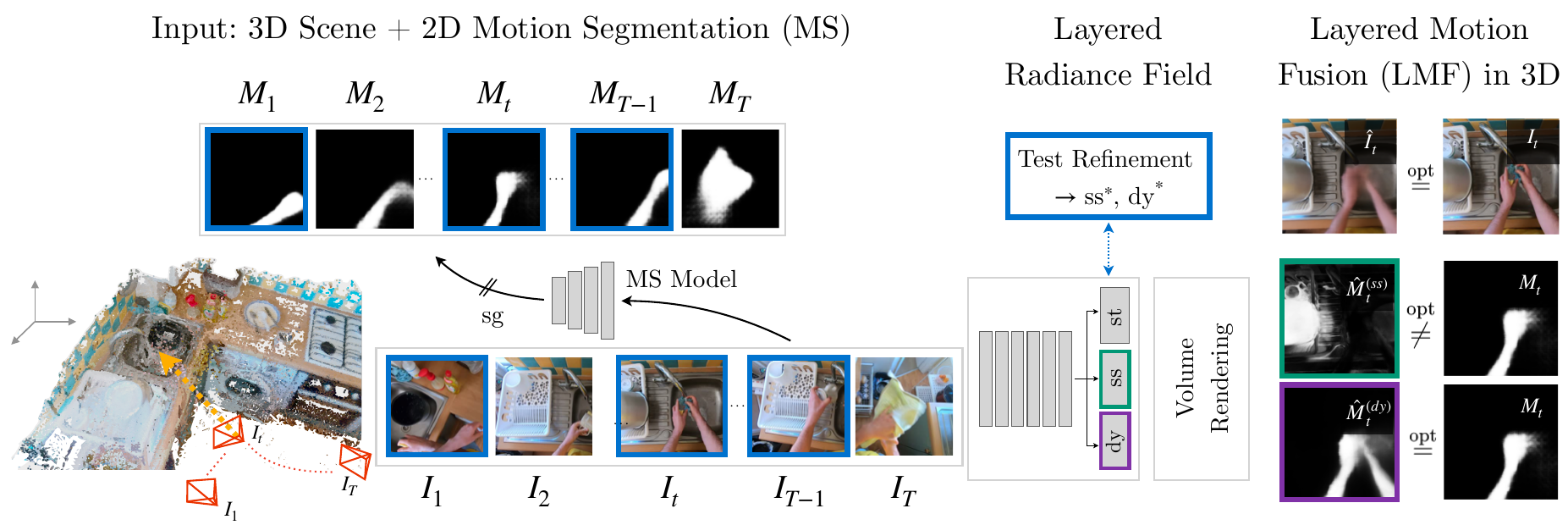}
\caption{\textbf{Overview of our method}. Given a layered radiance field with static, semi-static and dynamic layers, our method fuses pseudo-labels $M_t$ from a 2D segmentation method into its 3D representation. The static layer (st) is not updated as it does not learn any dynamics. The semi-static (ss) and dynamic (dy) layer  produce segmentation masks $\hat{M}_t^{(ss)}, \hat{M}_t^{(dy)}$, which are improved through Layered Motion Fusion (LMF) that consists of the RGB loss, the positive motion fusion and the negative motion fusion loss. The semi-static and dynamic models are updated to $\text{ss}^*$ and $\text{dy}^*$ through test-time refinement by focusing on frames that are selected for analysis (highlighted in blue).
}
\label{fig:method}
\end{figure*}

\newcommand{\x}{\boldsymbol{x}}
\newcommand{\y}{\boldsymbol{y}}
\newcommand{\z}{\boldsymbol{z}}
\newcommand{\w}{\boldsymbol{w}}
\newcommand{\bu}{\boldsymbol{u}}
\newcommand{\bnu}{\boldsymbol{\nu}}

Our method is given an egocentric video sequence $
\mathbf{V} = \{I_1, I_2, \dots, I_T\}$ with corresponding camera geometry, where each $I_t$ is a frame at time $t$ out of a total number of $T$ frames, and a 2D motion segmentation model $\mathcal{M}$ that outputs a sequence of motion segmentation masks $\mathbf{M} = \{M_1, M_2, \dots, M_T\}$ corresponding to the frames of the video. It then integrates the motion segmentation masks $
\mathbf{M}$ into a dynamic 3D model. The desired output is a set of enhanced 3D segmentation masks $\mathbf{\hat{M}} = \{\hat{M}_1, \hat{M}_2, \dots, \hat{M}_T\}$ that accurately represent the segmented dynamic and semi-static objects in the 3D space across the video sequence.\\
The following subsections first provide background on Neural Radiance Fields (NeRFs) in \cref{sec:background}, and then describe the integration of motion segmentation into layered NeRFs in \cref{s:motion-fusion} and how to boost their segmentation capability even further with test-time refinement in \cref{s:ttr}.

\subsection{Background on Neural Radiance Fields}%
\label{sec:background}

Let $I : \Omega \rightarrow \mathbb{R}^3$ be an image, where $\Omega \subset \mathbb{R}^2$ is the image domain (generally a rectangle).
Let $\bu = (u_x,u_y,1)$ be the homogeneous coordinate of a pixel, where $(u_x,u_y) \in \Omega$.
Let $\pi$ be a camera that images the 3D scene.
We define a ray as the parametric curve
$
\x_\tau = \x_0 - \bnu \tau
$,
where $\x_0$ is the center of projection (camera center), $\tau \in [0,\infty)$ is the distance traveled, and $\bnu$ is the direction of the ray, intersecting pixel $\bu$, where all the quantities are defined in the world reference frame system (rather than the camera's one).

A \emph{radiance field} is a pair of functions $(\sigma,c)$.
The opacity $\sigma : \mathbb{R}^3 \rightarrow \mathbb{R}_+$ maps 3D points to opacity values and the second function assigns a directional color $c : \mathbb{R}^3 \times \mathbb{S}^2 \rightarrow \mathbb{R}^3$ to each 3D point $\x$ and emission direction $\bnu$.
The color $I(\bu)$ of the image at pixel $\bu$ is given by the \emph{emission-absorption equation}
\begin{equation}\label{eq:ea}
I(\bu)
=
\int_{0}^\infty
c(\x_\tau, \bnu) \sigma(\x_\tau)
e^{-\int_0^\tau \sigma(\x_\mu)\,d\mu}\,d\tau.
\end{equation}

A \emph{dynamic} radiance field just adds an additional time variable $t\in[0,T]$ to these functions, so that
$
I : \Omega \times [0,T] \rightarrow \mathbb{R}^3
$,
$
\sigma : \mathbb{R}^3 \times [0,T] \rightarrow \mathbb{R}_+
$
and
$
c : \mathbb{R}^3 \times \mathbb{S}^2 \times [0,T] \rightarrow \mathbb{R}^3
$.

\subsection{Layered motion fusion (LMF)}%
\label{s:motion-fusion}

\paragraph{Layered neural radiance fields.}
Radiance fields have been originally designed for static scenes. In order to model dynamic videos, a natural extension assigns an individual radiance field per scene component as done in~\cite{martinbrualla2020nerfw,wu2022d,tschernezki21neuraldiff}. We refer to such models as \textit{layered neural radiance fields}. Since we are interested in the segmentation of dynamic objects in egocentric videos, we describe our method with respect to NeuralDiff~\cite{tschernezki21neuraldiff}, that was explicitly designed for such videos, but the fusion itself can easily be applied to other layered architectures as shown in the experiments (\Cref{sec:exp:analysis}). Assuming that we follow NeuralDiff, we decompose the scene into three layers: a static background layer, a semi-static layer, and a dynamic layer.
This corresponds to the functions
$
\sigma_\text{st}(\x), c_\text{st}(\bnu,\x)
$
for the static layer,
$
\sigma_\text{ss}(\x,t), c_\text{ss}(\bnu,\x,t)
$
for the semi-static one and
$
\sigma_\text{dy}(\bar \x,t), c_\text{dy}(\bar \bnu, \bar \x,t)
$
for the dynamic one, where only the last two layers are time-dependent.
The dynamic layer is defined w.r.t.~camera coordinates $\bar \x = \pi(\x)$ instead of world coordinates, where $\pi : \mathbb{R}^3\rightarrow\mathbb{R}^3$ is the world to camera coordinate transformation.
This is because %
in egocentric videos such as~\cite{tschernezki23epic}, the dynamic part is caused by the observer interacting with the world and so it is easier to model this part of the scene from their perspective.

The three fields (\ie layers) are combined into a single one via the equations
\begin{align}
\label{eq:sigma-combo}
\sigma(\x,t) = \sigma_\text{st}(\x) + \sigma_\text{ss}(\x,t) + \sigma_\text{dy}(\pi(\x),t)
\end{align}%
\begin{equation}
    \begin{multlined}
c(\x,\bnu) = \frac{
c_\text{st}(\bnu,\x)\sigma_\text{st}(\x) +
c_\text{ss}(\bnu,\x,t)\sigma_\text{ss}(\x,t)}{\sigma(\x,t)} + \\ \frac{
c_\text{dy}(\bar \x,\bar \bnu, t)\sigma_\text{dy}(\bar \x,t)}
{
\sigma(\x,t)
}.
\label{eq:c-combo}
\end{multlined}
\end{equation}

Note that colors are mixed and weighed by the opacity~\cite{tschernezki21neuraldiff}.
These equations can be used to render image $\hat I(\bu,t)$ as a function of time using~\Cref{eq:ea}.

On top of colors and opacities, we also predict an uncertainty value for each layer:
$
\beta_\text{st}(\x),
\beta_\text{ss}(\x,t), \text{and~}
\beta_\text{dy}(\bar \x,t) \in \mathbb{R}_+.
$
These are then projected to the image domain using \Cref{eq:sigma-combo,eq:ea} where $c$ is replaced by $\beta$ to render an uncertainty image $B(\bu,t)$.
This is used in a self-calibrated robust loss~\cite{novotny18capturing,kendall17what}:
\begin{equation}
\begin{multlined}
    \mathcal{L_\text{rgb}}(\hat I, I, B, t)
=
\frac{1}{|\Omega|}\sum_{\bu \in \Omega}
\frac{
\|\hat I(\bu,t) - I(\bu,t)\|^2
}
{
   2 B(\bu,t)^2
}
+ \\ \log B(\bu,t)^2
\end{multlined}
\label{eq:rgb_loss}
\end{equation}

\paragraph{Network architecture and information sharing.}
The other important characteristic
of layered radiance fields is how the different functions are implemented by a neural network and how the parameters are shared between the layers.
This is achieved by composing networks as follows:
\begin{align}
(\sigma_\text{st}(\x),c_\text{st}(\x,\bnu))
&=
\Phi_\text{st}(\Phi_0(\x), \bnu)
\label{eq:net-st}
\\
(\sigma_\text{ss}(\x,t), c_\text{ss}(\x,\bnu,t))
&=
\Phi_\text{ss}(\Phi_0(\x), \bnu, \z_t)
\label{eq:net-ss}
\\
(\sigma_\text{dy}(\bar \x,t), c_\text{dy}(\bar \x, \bar \bnu,t))
&=
\Phi_\text{dy}(\bar \x, \bar \bnu,\z_t).
\label{eq:net-dy}
\end{align}
Thus, the static and semi-static layers share the same spatial features $\Phi_0$.
The semi-static layer also takes as input the time $t$ encoded a time-dependent feature vector $\z_t \in \mathbb{R}^D$.
The dynamic layer does not share features since it is defined in a different reference frame, but does use the time encoding $\z_t$.
Taken together, the vectors $\z$ thus form a matrix $Z \in \mathbb{R}^{T \times D}$.
In order to ensure smoothness in the motion, the matrix $Z$ is decomposed as the product $Z = \tilde Z F$ where $F \in \mathbb{R}^{P \times D}$ is a fixed Fourier-like basis with $P \ll T$ and $\tilde Z \in \mathbb{R}^{T \times P}$ are learned coefficients.

\paragraph{Fusing motion into the semi-static and dynamic layer.}
Recall that our goal is to obtain a segmentation that separates dynamic and semi-static objects, which naturally emerges from the decomposition offered by layered neural fields.
However, experiments conducted in~\cite{tschernezki23epic} show that, while layered radiance fields improve the segmentation of semi-static components over off-the-shelf 2D motion segmentation methods~\cite{yang2021self}, they struggle with dynamic ones. Our idea is thus to use 3D reconstruction as a \emph{fusion} network, in line with \cite{zhi21in-place,tschernezki22neural,kobayashi2022distilledfeaturefields,bhalgat23contrastive}.
The key difference is that our scene is in motion
instead of being static. Additionally, %
fusion involves two different layers, representing the semi-static and dynamic motion.

\newcommand{\p}{\boldsymbol{p}}

We consider a motion segmentation algorithm that takes as input a video and outputs a segmentation mask $M(\bu,t) \in [0,1]$ for each frame, where $0$ means background and $1$ means foreground. We use these sparse, noisy and partial labels and fuse them into a joint implicit 3D space. This in turn enables the method to render \textit{denoised} labels back to frames through its learned representation. We render the fused labels through masks obtained from the dynamic and semi-static layers as follows. We associate pseudo-colors to both layers with ${\p}_\text{ss} = (0, 1, 0)$ and ${\p}_\text{dy} = (0, 0, 1)$ in order to calculate the pixel-based (rendered) mask values $\hat{M}_\text{ss}(\bu, t)$ and $\hat{M}_\text{dy}(\bu, t)$, and point-based mask values $m_\text{ss}(\x_\tau, t)$ and $m_\text{dy}(\x_\tau, t)$ respectively.
This results in the following volume rendering equation for \textit{negative} motion fusion as the semi-static layer is supposed to be dissimilar to purely dynamic motion (push it away):
\begin{equation}
    \begin{multlined}
\hat{M}_\text{ss}(\bu, t)
= \\
\int_{0}^\infty m_\text{ss}(\x_\tau, t)\sigma(\x_\tau, t)
e^{-\int_0^\tau \sigma(\x_\mu, t)\,d\mu}\,d\tau \,,
\label{eq:mask_render}
\end{multlined}
\end{equation}
with
\begin{equation}
\begin{multlined}
        m_\text{ss}(\x, t) =
\frac{
\p_{\text{ss}, 1}\sigma_\text{st}(\x) +
\p_{\text{ss}, 2}\sigma_\text{ss}(\x,t)
}
{
\sigma(\x,t)
} + \\
\frac{
\p_{\text{ss}, 3}\sigma_\text{dy}(\bar \x,t)
}
{
\sigma(\x,t)
} = \frac{
0 +
\sigma_\text{ss}(\x,t) +
0
}
{
\sigma(\x,t)
}\,.
\end{multlined}
\label{eq:mask_ss}
\end{equation}

For positive motion fusion, we similarly calculate $\hat{M}_\text{dy}$ with \Cref{eq:mask_render} where $\p_\text{dy}$ is used for $m_\text{dy}$ with

\begin{equation}
    m_\text{dy}(\x, t) = \frac{
\sigma_\text{dy}(\bar{\x},t)
}
{
\sigma(\x,t)
}.
\label{eq:mask_dy}
\end{equation}

With the rendered masks from the semi-static and dynamic layers, we calculate the positive motion fusion (PMF)  and the negative motion fusion (NMF) losses respectively. Compared to NMF, the PMF loss pulls dynamic motion of the 2D model to the dynamic layer and is defined as

\begin{equation}
    \begin{multlined}
\mathcal{L}_\text{PMF}(\hat{M}_\text{dy}, M, t) = \\ \lambda_{\text{PMF}} \frac{1}{|\Omega|}\sum_{\bu \in \Omega}
\|\hat{M}_\text{dy}(\bu,t) - M(\bu,t)\|^2.
\label{eq:pos_loss}
\end{multlined}
\end{equation}

The NMF loss penalizes the semi-static (ss) model when the predicted mask \(\hat{M}_\text{ss}(\mathbf{u},t)\) deviates from the target value of 0 for the set of pixels $\mathbf{u} \in \Omega$. This function is weighted by a factor \(\lambda_{\text{NMF}}\), adjusting the emphasis on the penalty for incorrect predictions over these \textit{negative} samples. Furthermore, we binarize the mask from the motion segmentation model to $\bar{M} \in \{0, 1\}$ and use it to select the pixels that are dynamic with $\bar{\Omega} = \{u \in \Omega\ | \bar{M}(u) = 1$\}, resulting in the loss:
\begin{equation}
\mathcal{L}_\text{NMF}(\hat{M}_\text{ss}, M, t) =
\lambda_{\text{NMF}} \frac{1}{|\bar{\Omega}|}\sum_{\bu \in \bar{\Omega}}
\|\hat{M}_\text{ss}(\bu,t)\|^2.
\label{eq:neg_loss}
\end{equation}

The final loss is then $\mathcal{L} = \mathcal{L_\text{rgb} + \mathcal{L}_\text{PMF} + \mathcal{L}_\text{NMF}},$ as defined in \Cref{eq:rgb_loss,eq:pos_loss,eq:neg_loss}.

\subsection{Test-time refinement}
\label{s:ttr}

We follow \cite{izquierdo2023sfm,chen2019self} and optimize the model at test time over a set of specified frames $\mathcal{T}$ \textit{before} rendering the masks $\hat{M}_\text{ss}$ and $\hat{M}_\text{dy}$. We refer to this procedure throughout the paper as TR (test refinement).
In comparison to the more typical refinement setting, we only refine the semi-static and dynamic model $\Phi_{st}$ and $\Phi_{dy}$.
The rationale for that is the independence of the static model from the task of motion segmentation, \ie it does not capture any motion by design. Let $W_\text{st}$, $W_\text{ss}$ and $W_\text{dy}$ be the sets of parameters of the static, semi-static and dynamic models respectively. We obtain the parameters $W_\text{ss}^{*}$ and $W_\text{dy}^{*}$ of the refined models $\Phi_{ss}^{*}$ and $\Phi_{dy}^{*}$ with
\begin{equation}
\begin{multlined}
    (W_\text{ss}^{*}, W_\text{dy}^{*}) = \\ \arg\min_{W_\text{ss}, W_\text{dy}} \sum_{t \in \mathcal{T}} \mathcal{L}(W_\text{st}, W_\text{ss}, W_\text{dy}; I_t, M_t, t).
\end{multlined}
\end{equation}
The masks from \Cref{eq:mask_dy,eq:mask_ss} are rendered as described in \Cref{eq:mask_render} with the densities and colors of the static model from \Cref{eq:net-st}, and outputs from the refined semi-static and dynamic model with
\begin{align}
(\sigma_\text{ss}(\x,t), c_\text{ss}(\x,\bnu,t))
&=
\Phi_\text{ss}^{*}(\Phi_0(\x), \bnu, \z_t)
\\
(\sigma_\text{dy}(\bar \x,t), c_\text{dy}(\bar \x, \bar \bnu,t))
&=
\Phi_\text{dy}^{*}(\bar \x, \bar \bnu,\z_t).
\end{align}

Besides selecting only reference frames, we also explore sampling additional frames within the vicinity of reference frames, to see if the added temporal context further facilitates the refinement. %
Formally, given a set of frames $\mathcal{T} = \{t_1, t_2, \ldots, t_M\}$ with $t_i \in [0, T], \forall i \in \{1, 2, \ldots, M\}$, we define a set of neighboring frames for each test frame $t_i$ within a window $N$. The combined set of test frames and their neighboring frames is denoted as $\mathcal{T}_{N}$. For each test frame $t_i \in \mathcal{T}$, the neighboring frames within a window $N$ are defined as:
\begin{align}\label{eq:sample_nb}
\mathcal{N}(t_i, N) = \{ t_{i - N}, \ldots, t_{i - 1}, t_i, t_{i + 1}, \ldots, t_{i + N} \}.
\end{align}
The set of frames used for refinement, including both the test frames and their neighbors, is given by:
\begin{align}\label{eq:refine_sampled}
\mathcal{T}_{N} = \bigcup_{t_i \in \mathcal{T}} \mathcal{N}(t_i, N).
\end{align}

\section{Experiments}
\label{sec:exp}

\begin{figure*}[!ht]
\centering
\includegraphics[width=0.95\textwidth]{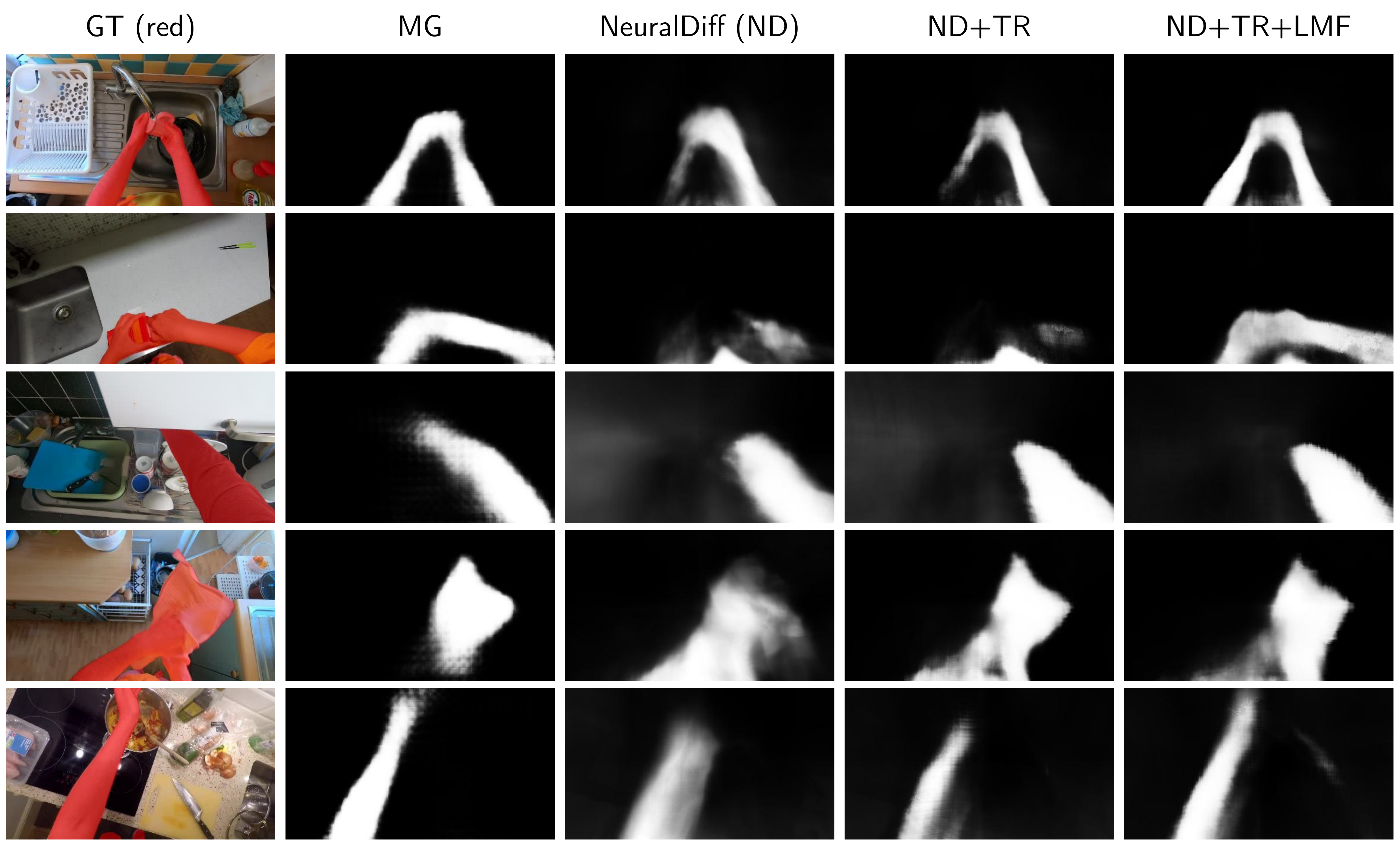}
\caption{\textbf{Qualitative results.}
The segmentations are produced by Motion Grouping (MG), NeuralDiff (ND), ND + Test-Time Refinement (TR), and ND + TR + Layered Motion Fusion (LMF). This shows the clear benefits of applying TR to ND, resulting in sharper segments such as in row 3, 4 and 5. The segmentation can be improved even further through LMF as shown in row 1 and 2.}
\label{fig:qualitative}
\end{figure*}

\begin{table}[t]
\caption{\textbf{Comparison with the state of the art}. Mean average precision~(mAP) on segmenting the dynamic (Dyn) and semi-static (SS) components and their union (SS+Dyn) for the UDOS 
task from EPIC Fields \cite{tschernezki23epic}. We report the results for our method (TR + LMF) combined with NeuralDiff (improvements \wrt ND shown in brackets). The original 3D baselines do not use any 2D fusion. %
Our approach enhances the segmentation of semi-static objects as a secondary benefit.
}\label{tab:sota}
\small
\centering
\scalebox{1.0}{%
\setlength{\tabcolsep}{3pt}
\begin{tabular}{lcc@{\hskip 8pt}ccc} %
\toprule
\textbf{Method} & \textbf{3D} & \textbf{2D} & \textbf{Dyn} & \textbf{SS} & \textbf{Dyn+SS}\\ 
\midrule
MG~{\cite{yang2021self}}     & \xmark & \cmark & 64.27 & 12.78 & 55.53 \\
NeRF-W~{\cite{martinbrualla2020nerfw}}     & \cmark & \xmark & 28.52 & 20.97 & 45.62             \\
NeRF-T~{\cite{gao2022monocular}}        &   \cmark & \xmark & 44.27 & 24.48 & 64.91 \\
NeuralDiff (ND)~{\cite{tschernezki21neuraldiff}}  & \cmark & \xmark & 55.58 & 25.55 & 69.74 \\ 
ND + TR + LMF (ours) & \cmark & \cmark & \textbf{72.51} & \textbf{27.70} & \textbf{74.21} \\
\midrule
&  & & +30.5\% & +8.4\% & +6.4\% \\
\bottomrule
\end{tabular}
}
\vspace{-0.1cm}
\end{table}

In the following, we will first describe the experimental details, and then compare our proposed method with the state of the art. The last sections contain an analysis of our method such as its application to other 3D methods and an ablation study.

\subsection{Experimental details}
Our experiments follow the \textit{Unsupervised Dynamic Object Segmentation} (UDOS) benchmark from the EPIC Fields dataset~\cite{tschernezki23epic}. EPIC Fields augments the EPIC-KITCHENS dataset~\cite{Damen2022RESCALING} with 3D camera information and use the provided segmentation masks of dynamic and semi-static objects for evaluation. We use the provided evaluation script and the pre-trained models (NeuralDiff, NeRF-W and T-NeRF -- referred to NeRF-T in our paper) for our experiments. The pre-trained models are trained for 20 epochs with a learning rate of $5 \times 10 ^{4}$ and cosine annealing schedule with an NVIDIA RTX A4000 per experiment. We set the parameters $\lambda_\text{PMF}$ and $\lambda_\text{NMF}$ of the LMF loss to 1.1 and 1.0 respectively. The fine-tuning takes about 22 minutes for 100 frames (about 13 seconds per frame). The rendering of a frame without fine-tuning takes about 5 seconds. For a fair comparison with the results from EPIC Fields, we %
use exactly the same frames as %
used to train their models. For motion fusion, we extract labels from the 2D motion segmentation model \textit{Motion Grouping} (MG)~\cite{yang2021self} that is used as 2D baseline in EPIC Fields. Further details such as network architecture or the training setup of MG can be found in the supplementary material of EPIC Fields \cite{tschernezki23epic}.

\subsection{Comparison with the state of the art}

We evaluate our method on the \textit{Unsupervised Dynamic Object Segmentation} (UDOS) task from EPIC Fields~\cite{tschernezki23epic} and compare to the 3D baselines NeuralDiff~\cite{tschernezki21neuraldiff}, NeRF-W~\cite{martinbrualla2020nerfw} and NeRF-T~\cite{gao2022monocular}. The results are shown in \Cref{tab:sota}. We observe improvements of up to 30\%. A positive byproduct of our method is the improvement of the semi-static segmentation, by up to 8\% in comparison to NeuralDiff. The joint segmentation of dynamic and semi-static objects improves by up to 6.4\%. The most important result is the improvement of NeuralDiff as the 3D baseline over MG as the 2D baseline, which was posed as an open problem and question in \cite{tschernezki23epic}. Our method outperforms MG by up to 8.2 mAP, while MG outperformed NeuralDiff previously by about 8.7 mAP.
Qualitative results comparing improvements over NeuralDiff and MG are shown in \Cref{fig:qualitative}.
Results that highlight the improvements of the semi-static segmentation are shown in the supplementary material in \Cref{fig:qualitative-ss}.

\begin{table*}
\caption{\textbf{Application to different 3D baselines}. We report the mean average precision~(mAP) on segmenting the dynamic (Dyn) and semi-static (SS) components of the scene, and also their union (SS+Dyn) for the UDOS task from EPIC Fields \cite{tschernezki23epic}. We apply a variant of our method to NeRF-W, NeRF-T and NeuralDiff. We observe consistent gains across all architectures. The relative improvements are shown in brackets.}\label{tab:quant_seg}
\small
\centering
\scalebox{1.0}{%
\setlength{\tabcolsep}{6pt}
\begin{tabular}
{l@{\hskip 80pt}cc@{\hskip 50pt}c@{\hskip 30pt}c@{\hskip 30pt}c}
\toprule
\textbf{Method} & \textbf{3D} & \textbf{2D} & \textbf{Dyn} & \textbf{SS} & \textbf{Dyn+SS}\\ 
\midrule
NeRF-W~{\cite{martinbrualla2020nerfw}}     & \cmark & \xmark & 28.52 & 20.97 & 45.62 \\
NeRF-W + TR + PMF & \cmark & \cmark & 34.20 (19.9\%) & 19.88 (-5.2\%) & 47.37 (3.8\%)             \\
\midrule
NeRF-T~{\cite{gao2022monocular}}        &   \cmark & \xmark & 44.27 & 24.48 & 64.91 \\
NeRF-T + TR + PMF &   \cmark & \cmark & 51.11 (15.4\%) & 23.24 (-5.1\%) & 68.87 (6.1\%)\\
\midrule
NeuralDiff ~{\cite{tschernezki21neuraldiff}}  & \cmark & \xmark & 55.58 & 25.55 & 69.74 \\ 
NeuralDiff + TR + PMF & \cmark & \cmark & \textbf{67.23} (20.9\%) & \textbf{26.61} (4.1\%) & \textbf{72.53} (4.0\%) \\
\bottomrule
\end{tabular}
}
\end{table*}

\begin{table}
\caption{\textbf{Effect of the temporal context on test refinement.} Mean average precision~(mAP) on segmenting the dynamic (Dyn) and semi-static (SS) components and their union (SS+Dyn) for the unsupervised dynamic object segmentation (UDOS)
task from EPIC Fields \cite{tschernezki23epic}, on a subset of 5 scenes and for 4 different numbers of neighbouring frames $N$ as defined in \cref{eq:sample_nb}. We set $N=0$ for the default setting without any neighboring frames, and compare to sampling 2, 5 and 20 frames. The relative improvements are shown in brackets.}\label{tab:ablation_temp_context}
\small
\centering
\scalebox{0.98}{%
\setlength{\tabcolsep}{2pt}
\begin{tabular}{lcc@{\hskip 2pt}ccc} %
\toprule
\textbf{Method} & \textbf{Dyn} & \textbf{SS} & \textbf{Dyn+SS}\\ 
\midrule
{NeuralDiff (ND)~{\cite{tschernezki21neuraldiff}}} & 58.12 & 25.84 & 70.49 \\ 
\midrule
ND + TR & \textbf{64.29} (10.6\%) & \textbf{26.26} (1.6\%) & \textbf{72.07} (2.2\%) \\
ND + TR-2 & 64.23 (10.5\%) & 26.23 (1.5\%) & 72.02 (2.2\%) \\
ND + TR-5 & 63.14 (8.6\%) & 26.21 (1.4\%) & 71.98 (2.1\%) \\
ND + TR-20 & 61.64 (6.1\%) & 25.95 (0.4\%) & 71.45 (1.4\%) \\

\bottomrule
\end{tabular}
}
\end{table}

Additionally, we compare our approach to a video object segmentation method specifically tailored for egocentric videos, focusing on the task of fine-grained hand-object segmentation. This comparison highlights the potential of 3D computer vision techniques to improve the performance of 2D video understanding methods. For this purpose, we select the state-of-the-art method EgoHOS~\cite{zhang2022fine} and report the results in \Cref{tab:method_egohos}. We observe that applying our method to MG \cite{yang2021self} results in a performance that is slightly better than EgoHOS. This result is significant, since EgoHOS requires supervision, while applying our method to MG works without any supervision. In addition we show that we can boost the performance of EgoHOS even further by combining it with our method. These results highlight the effectiveness and versatility of the proposed approach.

\begin{table}[ht]
\centering
\caption{\textbf{%
Hand-object segmentation in the dynamic setting.} We combine our method with the state-of-the-art hand-object segmentation method EgoHOS~{\cite{zhang2022fine}}. Our method with MG \cite{yang2021self} requires no supervision and is slightly better than EgoHOS, which is trained with supervision. Applying our method to EgoHOS improves it even %
further.}
\label{tab:method_egohos}
\small
\centering
\begin{tabular}{l@{\hskip 40pt}cccc}
\toprule
\textbf{Method} & \textbf{3D} & \textbf{2D} & \textbf{Supervision} & \textbf{mAP} \\
\midrule
MG~{\cite{yang2021self}}              & \xmark      & \cmark      & \xmark              & 64.27                \\
MG + Ours       & \cmark      & \cmark      & \xmark              & 72.51                \\
EgoHOS~{\cite{zhang2022fine}}          & \xmark      & \cmark      & \cmark              & 71.20                \\
EgoHOS + Ours   & \cmark      & \cmark      & \cmark              & \textbf{77.31}                \\ \bottomrule
\end{tabular}
\end{table}

\subsection{Analysis of model components}
\label{sec:exp:analysis}

\paragraph{Application to other architectures.} To show that our contributions can be applied beyond NeuralDiff, we also experiment with the architectures NeRF-T and NeRF-W (NeRF plus time~\cite{gao2022monocular} and NeRF in the wild~\cite{martinbrualla2020nerfw}).
While NeRF-W contains already a dynamic layer, we extend NeRF-T with the same dynamic layer as we use for our method as described in \Cref{sec:method}.
Furthermore, in NeRF-T, time is encoded using a positional encoding, whereas in NeRF-W, time is encoded by specifying and learning a separate latent vector $\z_t$ for each frame (similar to setting $P=D$ in the representation of $Z$).

We analyze the generalization of our method across these architectures in \Cref{tab:quant_seg}. Since NeRF-T and NeRF-W contain static and dynamic layers only, we can apply PMF, but omit NMF as it depends on a semi-static layer.
We observe that a combination of TR and PMF boosts the performance of all architectures in terms of the segmentation of dynamic objects. The decrease in performance of the semi-static prediction of NeRF-T and NeRF-W is expected as both architectures use only one layer to predict both dynamic and semi-static objects. Adding a semi-static layer to both models as shown in \Cref{tab:3d_baselines} of the supplementary results in improved semi-static performance.

\paragraph{Effect of temporal context on test refinement.} We defined the test refinement with respect to a number of neighboring frames in \Cref{eq:refine_sampled}. We analyze in \Cref{tab:ablation_temp_context} the influence of the temporal context on test-time refinement to find out if additional frames around test frames provide additional guidance to the model. We observe no benefit from adding frames.
This highlights the importance of enhancing the focus of the 3D model to achieve better segmentation.

\paragraph{Ablation study.} \Cref{tab:ablation_components} compares different combinations of components with NeuralDiff. We observe that motion fusion results overall in the highest relative gains for segmenting dynamic objects. The best results are obtained with test time refinement and layered motion fusion (LMF = PMF + NMF), resulting in an improvement of 30.4\%. The semi-static segmentation also improves significantly in this case, by 8.4\%. We observe as well that a joint combination of PMF and NMF is better than any of those components on their own -- with and without TR.

\begin{table}[!ht]
\caption{\textbf{Ablation}. Mean average precision~(mAP) on the segmentation of dynamic (Dyn) and semi-static (SS) components and their union (SS+Dyn), for the UDOS task from EPIC Fields \cite{tschernezki23epic}. The relative improvements are w.r.t.\ NeuralDiff.
The best results are obtained with layered motion fusion (LMF = PMF + NMF) and test time refinement (TR), and show that the different components improve the segmentation additively.}\label{tab:ablation_components}
\small
\centering
\scalebox{0.95}{%
\setlength{\tabcolsep}{4pt}
\begin{tabular}{lcc@{\hskip 10pt}ccc} %
\toprule
\textbf{Method} & \textbf{3D} & \textbf{2D} & \textbf{Dyn} & \textbf{SS} & \textbf{Dyn+SS}\\ 
\midrule
NeuralDiff (ND)~{\cite{tschernezki21neuraldiff}}  & \cmark & \xmark & 55.58 & 25.55 & 69.74 \\ 
\midrule
ND + NMF & \cmark & \cmark & 63.73 & 27.14 & 72.69 \\ \vspace{3pt}
 &  &  & +14.7\% & +6.2\% & +4.2\% \\
ND + PMF & \cmark & \cmark & 63.38 & 26.33 & 72.10 \\ \vspace{3pt}
& &  & +14.0\% & +3.1\% & +3.4\% \\
ND + PMF + NMF & \cmark & \cmark & 66.14 & 26.44 & 72.29 \\
& & & +19.0\% & +3.5\% & +3.7\% \\
\midrule
ND + TR & \cmark & \xmark & 62.29 & 26.02 & 71.66 \\ \vspace{3pt}
&  & & +12.0\% & +1.8\% & +2.8\% \\
ND + TR + NMF & \cmark & \cmark & 69.79 & \textbf{27.94} & \textbf{75.57} \\ \vspace{3pt}
& & & 25.5\% & 9.3\% & 8.3\% \\
ND + TR + PMF & \cmark & \cmark & 67.23 & 26.61 & 72.53 \\
& & & +20.9\% & +4.1\% & +4.0\% \\
\midrule
ND + TR + PMF + NMF & \cmark & \cmark & \textbf{72.51} & 27.70 & 74.21 \\
& & & +30.4\% & +8.4\% & +6.4\% \\
\bottomrule
\end{tabular}
}
\end{table}

\section{Conclusion}
\label{sec:ccl}

In this paper, we analyzed the limitations of current 3D methods when applied to the task of unsupervised dynamic object segmentation of long egocentric videos, as observed in the EPIC Fields benchmark~\cite{tschernezki23epic}.
We address these limitations through two contributions.
First, we introduce \textit{layered motion fusion}, which lifts motion segmentation predictions from a 2D-based model into layered radiance fields. This approach includes positive motion fusion, which pulls the predictions of the dynamic layer closer to the segmentation of the 2D model, and negative motion fusion, which pushes the predictions of the semi-static layer away from the segmentation of the 2D model. Both losses work in synergy, resulting in significant performance gains. We demonstrate that these effects can be further leveraged through test-time refinement, enabling the final model to outperform the 2D model used for training by a large margin. The proposed method is straightforward to implement, effective, adaptable to various 3D architectures and can be combined with a state-of-the-art hand-object segmentation approach to boost its performance. We believe that these results will inspire further research on the use of 3D geometry for egocentric scene and video understanding.

\clearpage
{
    \small
    \bibliographystyle{ieeenat_fullname}
    \bibliography{references}
}

\clearpage
\maketitlesupplementary
\phantomsection
\setcounter{section}{0}
\renewcommand{\thesection}{\Alph{section}}
\setcounter{figure}{0}

The supplementary material is structured as follows. First, we show how our proposed method improves the semi-static segmentation through negative motion fusion (NMF) in \Cref{sec:supp_nmf}. Then we discuss the limitations of dynamic segmentation for egocentric videos and our attempt to address them through layered motion fusion (LMF) and test-time refinement (TR) in \Cref{sec:limitations}. Next, we further investigate the precision of the motion segmentation predictions and justify their fusion in \Cref{sec:precision_predictions} and \Cref{sec:supp_fusion} respectively. \Cref{sec:generalizability_extended} provides further results regarding the generalizability of our method, and \Cref{sec:runtime_analysis} analyses its runtime.
\Cref{sec:app:broad} briefly discusses broader impacts.

\begin{figure*}
\centering
\includegraphics[width=0.85\textwidth]{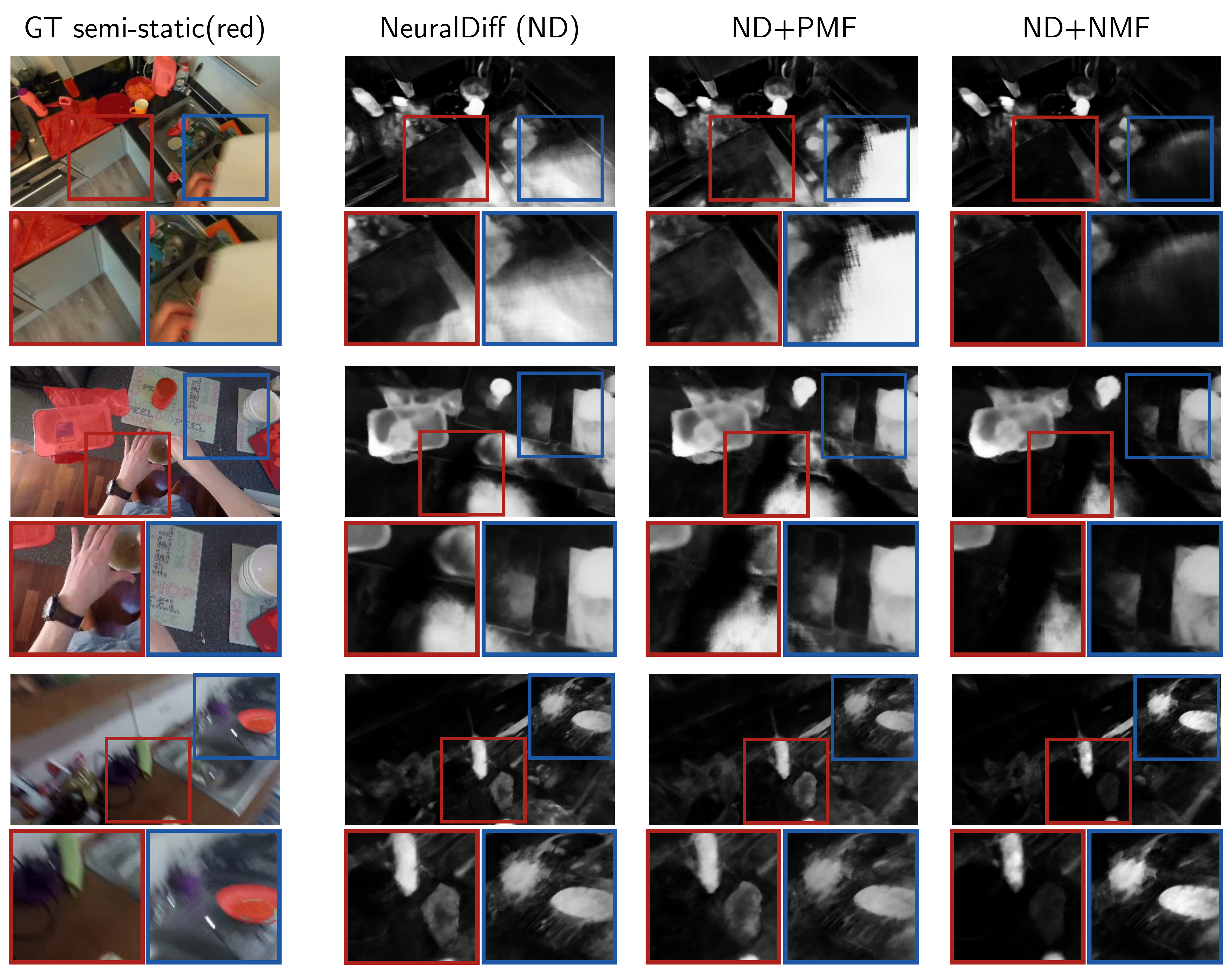}
\caption{\textbf{Qualitative semi-static results for motion fusion}.
The segmentations are produced by NeuralDiff (ND), ND + PMF, and ND + NMF. The positive motion fusion (PMF) loss does not prevent the semi-static layer from predicting dynamic objects.
In comparison, the negative motion fusion (NMF) loss removes artefacts from the semi-static predictions such as in row three.
It especially removes anything dynamic such as the cutting board in row one or the parts of the person and the bowl they are touching in row two.}
\label{fig:qualitative-ss}
\end{figure*}

\section{Improving semi-static segmentation}
\label{sec:supp_nmf}

We have shown in the main paper that, besides improving the dynamic segmentation, our method also improves the semi-static segmentation as well by making use of \textit{only dynamic} predictions of the 2D based segmentation model. We achieve this by preventing the semi-static model from predicting
anything that has a dynamic (pseudo-) label assigned. We highlight the benefits in \Cref{fig:qualitative-ss}. We see that NMF removes artefacts from the semi-static layer such as in row three. Additionally, NMF also reduces predictions of dynamic objects such as the cutting board in row one and the person in row two. As can be seen, such results cannot be achieved by training with PMF only, as this loss does not influence the semi-static layer directly.

\begin{figure*}
\centering
\includegraphics[width=1.0\textwidth]{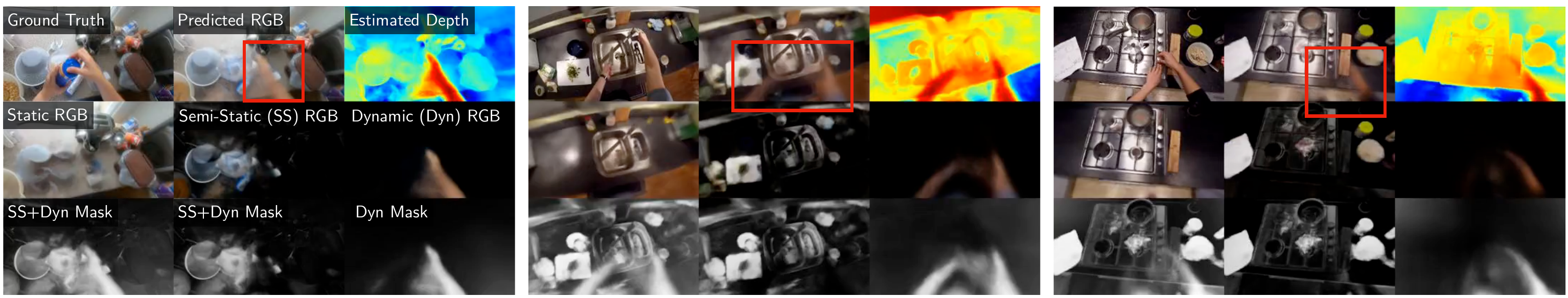}
\caption{\textbf{Missing geometry when segmenting dynamic objects}. While the model is able to segment the semi-static components of the scene, the dynamic one is rendered with less accuracy. This lack of geometric understanding of the dynamics of the scene hurts segmentation,
as the model can only segment objects whose geometry is captured.
}
\label{fig:limitations_rendering}
\end{figure*}

\begin{figure*}
\centering
\includegraphics[width=1.0\textwidth]{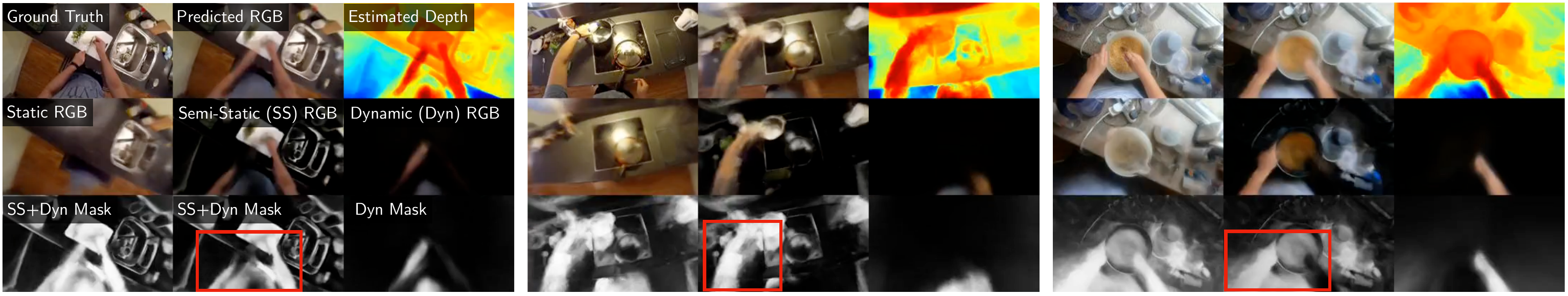}
\caption{\textbf{The semi-static layer captures dynamic objects incorrectly}. We observe that the model segments dynamic components (the person and the objects they are holding) incorrectly into the semi-static stream. This represents a significant limitation of dynamic neural rendering methods. It can be resolved with the proposed layered motion fusion that integrates the predictions of a 2D based segmentation into its 3D representation.}
\label{fig:limitations_segmentation}
\end{figure*}

\section{Limitations of dynamic segmentation}
\label{sec:limitations}

The authors of the EPIC Fields \cite{tschernezki23epic} dataset observed that the performance of state-of-the-art dynamic neural rendering methods strongly depends on the type of motion. Their results show a significant gap in the reconstruction quality between the dynamic and the static parts of videos, pointing to the current limitations when handling dynamic objects. 
This limitation also appears in the segmentation of dynamic objects: While the 3D-based models from EPIC Fields outperform MG as a 2D baseline on the semi-static setting, they all perform worse when used for the segmentation of dynamic objects. For example, they report a mean Average Precision (mAP) score of 55.58 for NeuralDiff, while MG achieves 64.27 -- a significant gap of about 15\%. 

We hypothesize that this difference is caused by two factors. First, we observe that the semi-static model sometimes captures dynamic objects as shown in \Cref{fig:limitations_segmentation}. The figure shows that the model reconstructs the scene well in comparison to \Cref{fig:limitations_rendering}, but it assigns the person and the object they are holding incorrectly to \textit{the semi-static layer}. To circumvent this problem, we make use of labels predicted from a 2D motion segmentation method, and fuse them into the 3D model. This procedure regularises both layers (semi-static and dynamic) of the model and helps specifically the dynamic layer to learn about moving objects. Second, the fusion of motion segmentation predictions can only be achieved if the geometry is captured correctly. 
This means that, for example, if the arms are not captured by the radiance field, it is impossible to fuse motion into them as their geometry is missing.
Examples of failure cases are shown in \Cref{fig:limitations_rendering}. The red rectangles highlight the reconstruction of dynamic objects and show a significant mismatch between the predicted RGB image and the ground truth. This in turn leads to a deterioration of the segmentation ability, since the 3D based model can only segment what it can reconstruct. We address this issue through test-time refinement.

Aside from the challenges posed by dynamic segmentation with geometry, the task also proves to be a significant hurdle for 2D-based \textit{supervised} methods, such as EgoHOS \cite{zhang2022fine}, as demonstrated in \cref{tab:method_egohos}. While this model is trained to segment hands and objects with specific supervision, it achieves a lower score compared to our method. More recent papers provide further evidence of the difficulty of segmenting dynamic objects in egocentric videos such as \cite{Bhalgat24b} and \cite{Plizzari2025OSNOM}.

\section{Precision of motion segmentation output}
\label{sec:precision_predictions}

\begin{figure*}
\centering
\includegraphics[width=0.95\textwidth]{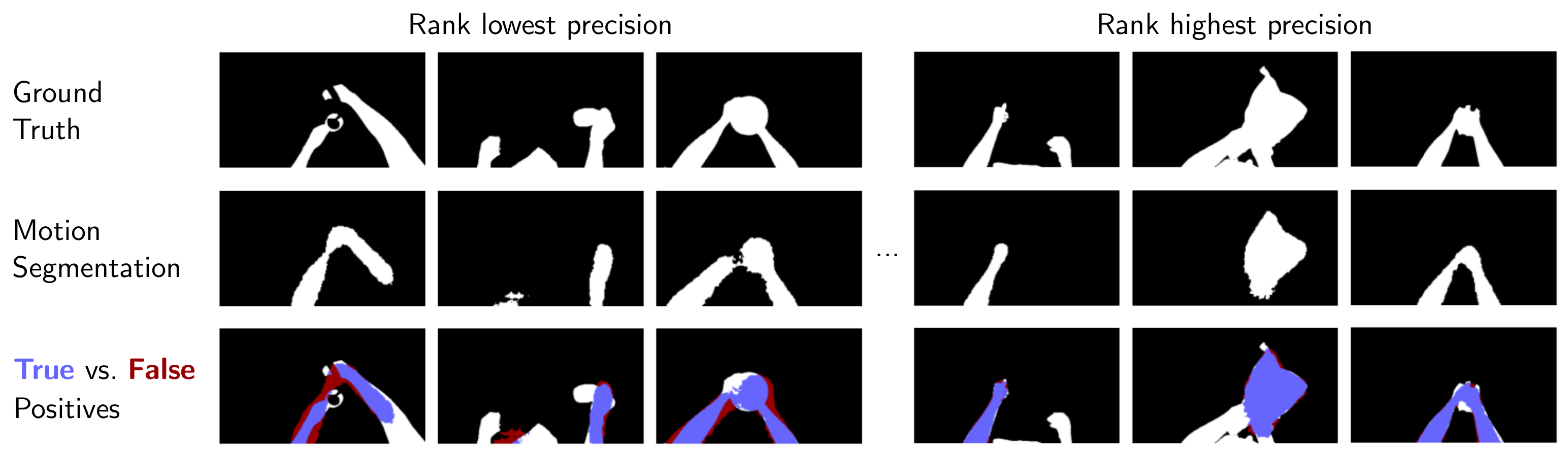}
\caption{\textbf{Precision of motion segmentation predictions}. The motion segmentation predictions have unbalanced error probabilities for egocentric videos. We observe that the motion segmentation model is less likely to produce false positives and has a high precision as shown in the ranking of samples with respect to their precision.
We rank segmentations with the lowest precision from right to left in the first three columns and those with the highest precision from left to right in the last three columns.
}
\label{fig:precision_rank}
\end{figure*}

\begin{figure}
\centering
\includegraphics[width=0.95\columnwidth]{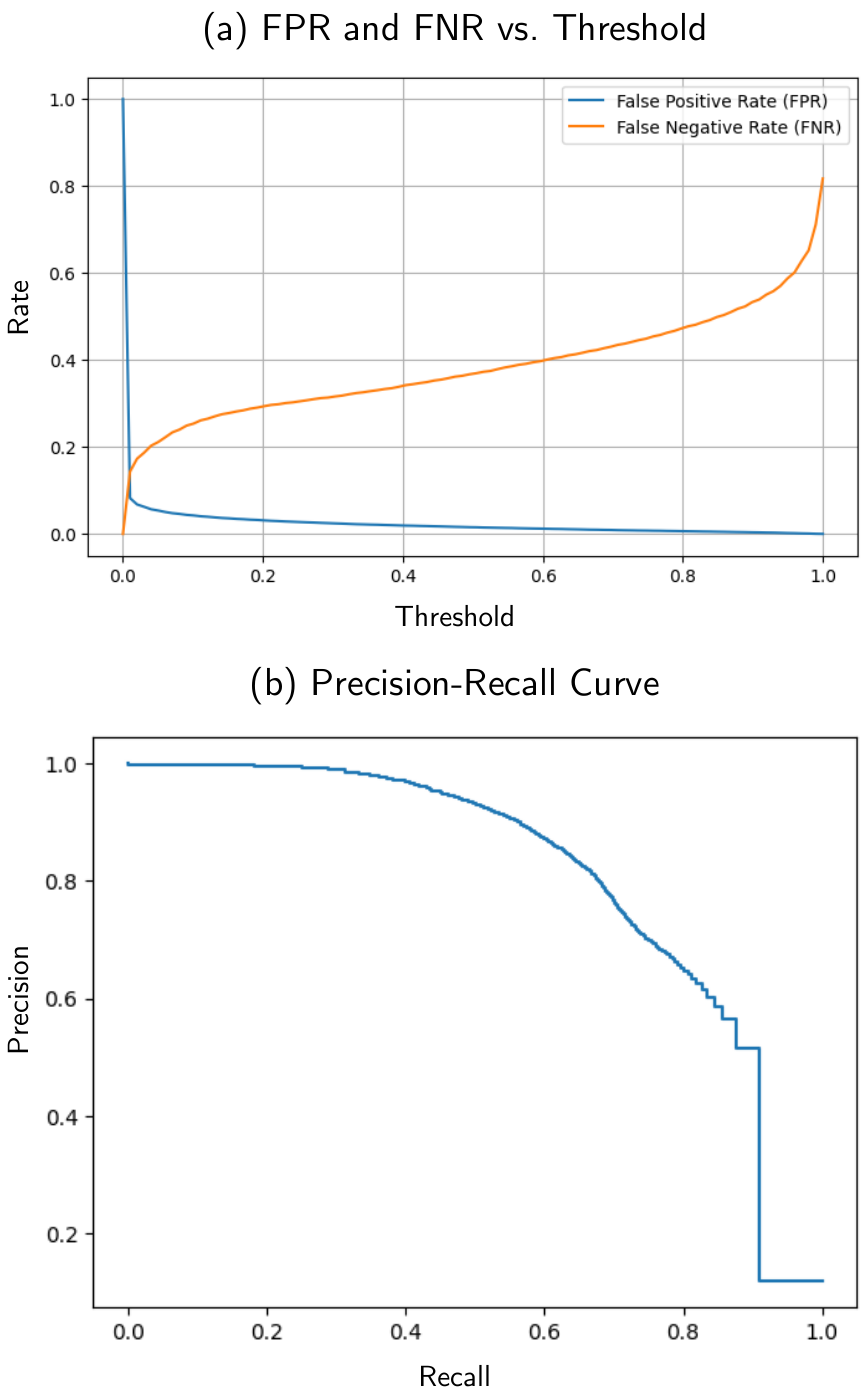}
\caption{\textbf{Analysis of true positives from motion segmentation predictions.} (a) We observe that the probabilities from the motion segmentation method are unbalanced in terms of the false positive rate (FPR) and false negative rate (FNR). The model is rarely predicting positives incorrectly for varying thresholds.  (b) The precision-recall curve inclines towards the top right and suggests that the motion segmentation method is highly effective in differentiating between the positive and negative classes.}
\label{fig:precision_plot}
\end{figure}

We observe that the motion segmentation predictions are unbalanced in terms of the false positive rate (FPR) and false negative rate (FNR). An increase in FPR indicates that the model is incorrectly labeling more negative instances as positive, which would decrease the precision since it is negatively impacted by an increase in FP (more false positive predictions reduce the fraction of true positive predictions among all positive predictions). While a high precision is generally desirable for the fusion of labels, the negative fusion loss benefits in particular from it, since the semi-static model can only learn to ignore dynamic objects that are actually positives. An analysis of the error rates can be found in \Cref{fig:precision_plot} and a qualitative visualization of the precision of the motion segmentation predictions can be found in \Cref{fig:precision_rank}.

\section{Fusion of motion segmentation predictions}
\label{sec:supp_fusion}

The work from \cite{zhi21in-place} has shown that noisy or sparse labels can be fused into 3D space through neural rendering in static scenes. They observe that the accuracy of the fusion decreases with a significant increase in noise and sparsity. For the case of fusing motion segmentation masks from a 2D-based model, such as MG \cite{yang2021self}, into a dynamic neural rendering representation, we therefore require labels with high precision. We analyze the FPR rate in \Cref{fig:precision_plot} and note that MG rarely predicts positives incorrectly for varying thresholds.
This observation is not only important for the fusion of labels into a single layer (similar to Semantic-NeRF~\cite{zhi21in-place}), but even more so when fusing them into two layers as proposed in our method. While the dynamic layer simply learns to predict the dynamic labels from the motion segmentation masks, the semi-static layer is penalized for predicting anything that should belong to the dynamic layer. As the semi-static layer learns to exclude the predictions from the 2D-based segmentation model, the dynamic layer is forced indirectly to predict them instead. This results in a higher overall confidence either for the semi-static model \textit{or} the dynamic model. In comparison, applying only one of the losses can result in predictions of the semi-static and dynamic layer that have equal confidence (probability of 0.5), as the segmentation itself depends on the rendering with multiple layers as defined in the rendering equations from \cite{tschernezki21neuraldiff}. Another positive side-effect of the layered motion fusion is that the semi-static model improves as well, as shown in the results, by making use of \textit{only dynamic} labels from the 2D-based model. Qualitative examples are visualized in \Cref{fig:qualitative-ss}.

\section{Generalizability of semi-static component}
\label{sec:generalizability_extended}

The methods NeRF-W \cite{martinbrualla2020nerfw} and NeRF-T \cite{gao22monocular} are not designed for segmenting semi-static objects, and therefore lack a semi-static layer.
This results in a lower performance, because we cannot apply NMF\@.
We chose the architectures from EPIC Fields \cite{tschernezki23epic} for a fair comparison, but extend them to show the generalizability of our method.
We add a semi-static layer similar to that used in NeuralDiff.
The results, reported in \Cref{tab:3d_baselines}, are extensions of the experiments from \Cref{tab:quant_seg} and \Cref{tab:ablation_components} in the main paper.
We observe that the semi-static layer improves the performance on its own. Adding NMF improves the performance of NeRF-W and NeRF-T for all types of motion (including semi-static) even further, similar to its application in NeuralDiff.

\begin{table}
\caption{\textbf{Application to different 3D baselines.}
We report the mean average precision~(mAP) on segmenting the dynamic (Dyn) and semi-static (SS) components of the scene, and also their union (SS+Dyn).
We modify NeRF-W and NeRF-T to a three-layer architecture (indicated by $^*$), which enables us to apply LMF (\ie PMF+NMF) as opposed to PMF only (\Cref{tab:quant_seg} from main paper).}\label{tab:3d_baselines}
\small
\centering
\scalebox{0.93}{%
\setlength{\tabcolsep}{4pt}
\begin{tabular}
{lccc}
\toprule
\textbf{Method} & \textbf{Dyn} & \textbf{SS} & \textbf{Dyn+SS}\\ 
\midrule
NeRF-W \cite{martinbrualla2020nerfw}     & 28.5 & 20.9 & 45.6 \\
+ TR + PMF & 34.2 (20.0\%) & 19.8 
(-5.3\%) & 47.3 (3.7\%) \\
+ TR + PMF $^*$ & 34.5 (21.1\%) & 21.1 (+1.0\%) & 48.2 (5.7\%) \\
+ TR + PMF + NMF $^*$ & 36.6 (28.4\%) & 21.6 (+3.3\%) & 49.4 (8.3\%) \\
\midrule
NeRF-T \cite{gao2022monocular}        & 44.2 & 24.4 & 64.9 \\
+ TR + PMF & 51.1 (15.6\%) & 23.2 (-4.9\%) & 68.8 (6.0\%)\\
+ TR + PMF $^*$ & 52.9 (19.7\%) & 24.7 (+1.2\%) & 69.2 (6.6\%)\\
+ TR + PMF + NMF $^*$ & 56.1 (26.9\%) & 25.5 (+4.5\%) & 69.9 (7.7\%)\\
\bottomrule
\end{tabular}
}
\vspace{-18pt}
\end{table}

\section{Runtime analysis}
\label{sec:runtime_analysis}

The fine-tuning takes about 22 minutes for 100 frames or about 13 seconds per frame. The rendering of a frame without fine-tuning takes about 5 seconds. Therefore, the required time for rendering would increase to 18 seconds with the fine-tuning. Given that, our method improves the mAP score by up to 30\%. The runtime could be significantly reduced to a fraction of its current value by adopting a more advanced architecture, such as Gaussian Splatting \cite{kerbl3Dgaussians}, though this would require a non-trivial adaptation.

\section{Broader impact}
\label{sec:app:broad}
This method, because of potential applications in augmented reality, could have some positive impact as it could be used within an AI assistant. 
There are also potential drawbacks to such technologies, since improved AR technologies could potentially be exploited for deceptive purposes.

\end{document}